\documentclass[sigconf, natbib=false, nonacm]{acmart}

\usepackage[datamodel=acmdatamodel, style=acmnumeric, sorting=none, backend=biber]{biblatex}
\usepackage{amsmath, amsfonts}

\usepackage{xspace}

\usepackage{orcidlink}

\usepackage{graphicx}
\usepackage{subcaption}

\usepackage[inline]{enumitem}
\setlist*[itemize]{labelindent=10pt, itemindent=0pt, leftmargin=*}

\usepackage{booktabs}
\usepackage{multirow}
\usepackage{tabularx}
\usepackage{colortbl}

\usepackage{pgfplots}
\pgfplotsset{compat=1.18}
\usepgfplotslibrary{statistics}

\usepackage[nameinlink]{cleveref}

\usepackage{csquotes}
\usepackage{textcomp}

\usepackage{balance}

\definecolor{greenCustom}{HTML}{EAF7EA}
\definecolor{yellowCustom}{HTML}{FFF8DC}
\definecolor{greyCustom}{HTML}{E6E6E6}
\definecolor{beigeCustom}{HTML}{FFEBCD}

\newcommand{\rowHLBaseline}{\rowcolor{greyCustom}}
\newcommand{\rowHLBest}{\rowcolor{beigeCustom}}
\newcommand{\posG}[1]{~{\scriptsize\textcolor{green!60!black}{(+#1\%)}}}
\newcommand{\negG}[1]{~{\scriptsize\textcolor{green!60!black}{(-#1\%)}}}
\newcommand{\posR}[1]{~{\scriptsize\textcolor{red!70!black}{(+#1\%)}}}
\newcommand{\negR}[1]{~{\scriptsize\textcolor{red!70!black}{(-#1\%)}}}
\newcommand{\negNum}[1]{~{\scriptsize\textcolor{red!70!black}{-#1}}}
\DeclareRobustCommand{\MergeOver}{{\sffamily\scshape MergeOver}\xspace}
\newcommand{\ditto}{\textquotedbl}

\begin{document}

\title[Post-Training Token Merging for Recursive ViTs]{\MergeOver: Post-Training Token Merging\protect\\ for Recursive Vision Transformers}

\author{Junseo {Kim}}
\orcid{0009-0008-4806-5379}
\affiliation{%
	\institution{Computer Architecture for Embedded Systems, University of Twente}
	\city{Enschede}
	\country{The Netherlands}}
\email{j.kim-5@student.utwente.nl}

\author{Uraz {Odyurt}}
\orcid{0000-0003-1094-0234}
\affiliation{%
	\institution{Faculty of Engineering Technology, University of Twente}
	\city{Enschede}
	\country{The Netherlands}}
\email{u.odyurt@utwente.nl}

\author{Amirreza {Yousefzadeh}}
\orcid{0000-0002-2967-5090}
\affiliation{%
	\institution{Computer Architecture for Embedded Systems, University of Twente}
	\city{Enschede}
	\country{The Netherlands}}
\email{a.yousefzadeh@utwente.nl}

\renewcommand{\shortauthors}{J. Kim et al.}

\begin{abstract}
Vision Transformers (ViTs) demonstrate exceptional performance in computer vision but suffer from large parameter counts and quadratic computational complexity, severely limiting their deployment on resource-constrained edge hardware. While recursive weight-sharing reduces parameter counts and token merging mitigates computational and memory bottlenecks, integrating these two paradigms without costly retraining is non-trivial, leaving this intersection largely unexplored. We propose \MergeOver, a post-training approach that integrates Token Merging (ToMe) into the recursively weight-shared Sliced Recursive Transformer (SReT). Through an \emph{Unmerge} tracking stack, constraint-safe merge-rate adjustment, and synchronised token-mass tracking across spatial permutations, \MergeOver resolves the spatial and merging constraints of this integration. We further employ a stage-wise single-shot schedule that performs token reduction at the first block of each stage and maintains a fixed sequence length throughout its subsequent recursive iterations. Benchmarked on ImageNet-1K, our selected configuration reduces top-1 accuracy by 1.47 percentage points. On the GPU, it reduces peak activation memory by 37.3\% and 38.4\% at batch sizes 1 and 16, while throughput decreases by 21.7\% at batch size 1 but increases by 21.7\% at batch size 16. On a Raspberry Pi 5 (ARM CPU), it reduces latency by 2.4\% and 17.6\% at batch sizes 1 and 16. These results show that \MergeOver can recover a meaningful part of the throughput and memory cost that recursive weight-sharing introduces, without retraining, and provides a baseline for combining token merging with hierarchical recursive transformers.
\end{abstract}

%


\maketitle


\section{Introduction}
\label{sec:introduction}
The self-attention mechanism of the Transformer architecture~\cite{transformer-vaswani2017} has revolutionised computer vision through the introduction of the Vision Transformer (ViT)~\cite{vit-dosovitskiy2021}. However, this architecture results in massive parameter counts and quadratic computational complexity, $\mathcal{O}(N^2)$, with respect to sequence length, $N$, limiting practical deployment on resource-constrained edge hardware~\cite{mobilevit-mehta2022, efficientvit-liu2023}. To reduce parameter counts, recursive weight-sharing paradigms have been explored, which repeatedly route the input sequence through a single transformer block, applying identical weights across multiple passes~\cite{ut-dehghani2019, minivit-zhang2022, sret-shen2022}. However, this approach results in increased computational complexity and Peak Activation Memory (PAM).

To explicitly target these computational and memory bottlenecks, token reduction paradigms have emerged, aiming to reduce the token sequence during inference. Modern strategies include token pruning~\cite{dynamicvit-rao2021, evit-liang2022}, token merging~\cite{tome-bolya2023, pitome-tran2024}, and token fusion~\cite{tofu-kim2024}. Amongst these, Token Merging (ToMe)~\cite{tome-bolya2023} stands out as a promising post-training framework. 

Recursive weight-sharing and token reduction have developed as independent lines of work, and their combination has not been studied. Moreover, while alternative efficient architectures exist, they rely on re-training phases, distillation, or structural redesigns~\cite{mobilevit-mehta2022, efficientvit-liu2023, fastvit-vasu2023, repvit-wang2024, madtp-cao2026}. This creates a literature gap for post-training compression methods. We address this gap by proposing a post-training compression approach that combines recursive weight-sharing with token merging. We validate this integration through a concrete case study that integrates the Sliced Recursive Transformer (SReT)~\cite{sret-shen2022} with ToMe~\cite{tome-bolya2023}. Our developed source code is openly available~\cite{Kim:2026:CODE}.

Naive integration introduces severe architectural friction. First, token reduction through merging violates the rigid 2D spatial grid required by SReT's hierarchical convolutional pooling layers. Second, unconstrained merging can violate the strict group-size boundaries inherent to the Sliced Group Self-Attention (SGA) of SReT. Third, SReT's spatial permutations break the tracking necessary for the proportional attention of ToMe. Resolving these incompatibilities is mandatory for a successful integration. 

\paragraph*{Contributions}
Specifically, we make the following contributions:
\begin{itemize}
    \item The formulation of \MergeOver, a post-training approach that enables token merging within a hierarchical recursive transformer, without retraining.
    \item An analysis of token reduction scheduling strategies, culminating at a stage-wise single-shot reduction strategy compatible with SReT.
    \item A cross-platform evaluation of \MergeOver against an uncompressed recursive baseline on the ImageNet-1K dataset~\cite{imagenet-deng2009}.
\end{itemize}

Following this introduction, \Cref{sec:background-related-work} contextualises our work within literature, while \Cref{sec:methodology-implementation} elaborates the SReT and ToMe integration. \Cref{sec:experimental-setup} details the experimental setup, complemented by \Cref{sec:results} covering benchmarking results and discussion. \Cref{sec:future-work,sec:conclusion} point out the limitations and provide our concluding remarks.

\section{Background and related work}
\label{sec:background-related-work}
We review the existing literature and foundational concepts related to \MergeOver. 

\subsection{Vision Transformers and spatial pyramids}
Modern ViTs stem from the Transformer architecture~\cite{transformer-vaswani2017}, computing global dependencies via scaled dot-product attention over Queries ($Q$), Keys ($K$), and Values ($V$):
\begin{equation}
    \operatorname{Attention}(Q, K, V)
    =
    \operatorname{Softmax}\left(
    \frac{QK^\top}{\sqrt{d_k}}
    \right)V,
    \label{eq:attention}
\end{equation}
where $d_k$ is the scaling dimension. ViT~\cite{vit-dosovitskiy2021} adapted this to computer vision by flattening 2D images into 1D patch sequences. However, standard ViTs maintain a strictly isotropic layout where sequence length and channel capacity remain uniform, resulting in a quadratic computational complexity relative to sequence length.

To mitigate this bottleneck, hierarchical spatial pyramids introduce progressive downsampling. While Swin~\cite{swin-liu2021} and PVT~\cite{pvt-wang2021} employ localised attention or spatial reduction, PiT~\cite{pit-heo2021} directly integrates CNN-based convolutional pooling. This effectively models complex spatial hierarchies, establishing the exact architectural blueprint inherited by recursive variants such as SReT~\cite{sret-shen2022}.

\subsection{Recursive weight-sharing}
Weight-sharing minimises parameters by reusing weights via recurrence. Pioneered by the Universal Transformer~\cite{ut-dehghani2019}, this paradigm limits parameter counts while allowing arbitrary computational depth. Early vision adaptations like MiniViT~\cite{minivit-zhang2022} applied layer-wise weight-multiplexing to simulate depth, but lacked true parametric recursion.

Pure spatial recurrence within a hierarchical architecture was realised by SReT~\cite{sret-shen2022}. Built on PiT's~\cite{pit-heo2021} convolutional pyramid blueprint, SReT routes sequences recursively through identical blocks with frozen parameters across consecutive loops. 

While SReT is highly parameter-efficient, repeating self-attention creates a severe computational bottleneck, and routing tokens through a shared parameter space accelerates feature densification. This severely penalises throughput and increases PAM, necessitating sequence-level compression.

\subsection{Token reduction}
To alleviate the quadratic complexity of ViTs, token reduction paradigms minimise spatial redundancy during inference via token pruning~\cite{dynamicvit-rao2021, evit-liang2022}, token merging~\cite{tome-bolya2023, pitome-tran2024}, or token fusion~\cite{tofu-kim2024}.

Token pruning permanently discards patches. However, methods like DynamicViT~\cite{dynamicvit-rao2021} require retraining, while post-training alternatives like EViT~\cite{evit-liang2022} rely heavily on the \emph{extra class token (CLS)}, which is incompatible with SReT's~\cite{sret-shen2022} global average pooling. Crucially, permanent deletion prevents reconstructing the rigid 2D grid required for inter-stage convolutional pooling.

In contrast, ToMe~\cite{tome-bolya2023} is a post-training module that merges similar tokens via Bipartite Soft Matching (BSM). To preserve proportional attention after merging, ToMe tracks the number of original patches represented by each token using a token-mass vector ($s$). The logarithm of the key-token masses is added to the attention logits before the softmax:
\begin{equation}
    \operatorname{Attention}(Q, K, V)
    =
    \operatorname{Softmax}\left(
    \frac{QK^\top}{\sqrt{d_k}} + \log(s_K)
    \right)V.
    \label{eq:proportional-attention}
\end{equation}

While variants like PiToMe~\cite{pitome-tran2024} optimise matching, ToMe's reversibility is uniquely valuable for hierarchical networks. Along with a merge operation, BSM enables an \emph{Unmerge} operation that duplicates features to restore original sequence length. This infrastructure provides the exact mechanism to satisfy the spatial layout constraints of SReT's inter-stage convolutional pooling.

Token fusion frameworks like ToFu~\cite{tofu-kim2024} hybridise these paradigms. However, combining destructive pruning with irreversible merging destroys bi-directional tracking. Lacking a structural Unmerge function, fusion remains incompatible with SReT's hierarchical pyramid.

\subsection{Alternative efficiency paradigms}
Beyond recursive weight-sharing and token reduction, several alternative paradigms target structural efficiency. Architectural redesigns (MobileViT~\cite{mobilevit-mehta2022}, EfficientViT~\cite{efficientvit-liu2023}), reparameterisations (FastViT~\cite{fastvit-vasu2023}, RepViT~\cite{repvit-wang2024}), and structured pruning (X-Pruner~\cite{xpruner-yu2023}) increase throughput by modifying layouts or cutting channels. However, their reliance on intensive retraining prevents post-hoc application to pre-trained models.

Conversely, implementation-level and precision-reduction frameworks optimise hardware execution without altering architecture. Hardware-aware kernels like FlashAttention~\cite{flashattention-dao2022} bypass memory bottlenecks via GPU SRAM tiling, but leave spatial token redundancy untouched. Similarly, post-training quantisation (FQ-ViT~\cite{fqvit-lin2022}) compresses weights and activations into low-bit formats during inference. By optimising memory access and numerical precision rather than parameter counts or sequence lengths, these approaches represent entirely complementary mechanisms that can integrate alongside our proposed approach.

\subsection{Multi-axis compression}
ViT optimisation has shifted from isolated, single-axis methods toward multi-axis joint compression. For instance, MADTP++~\cite{madtp-cao2026} unifies structural weight pruning and dynamic token pruning but requires retraining, preventing inference-only deployment. To bypass this overhead, post-training strategies like QUOTA~\cite{joint-li2026} merge low-bit quantisation with deterministic sequence pruning.

While highly efficient for isotropic ViTs, these paradigms rely exclusively on destructive, non-reversible operations. Crucially, the intersection of recursive weight-sharing and dynamic token merging remains unexplored. \MergeOver fills this void using a post-training integration of these two axes.

\section{Methodology and implementation}
\label{sec:methodology-implementation}
\begin{figure*}[htbp]
    \centering
    \includegraphics[width=\textwidth]{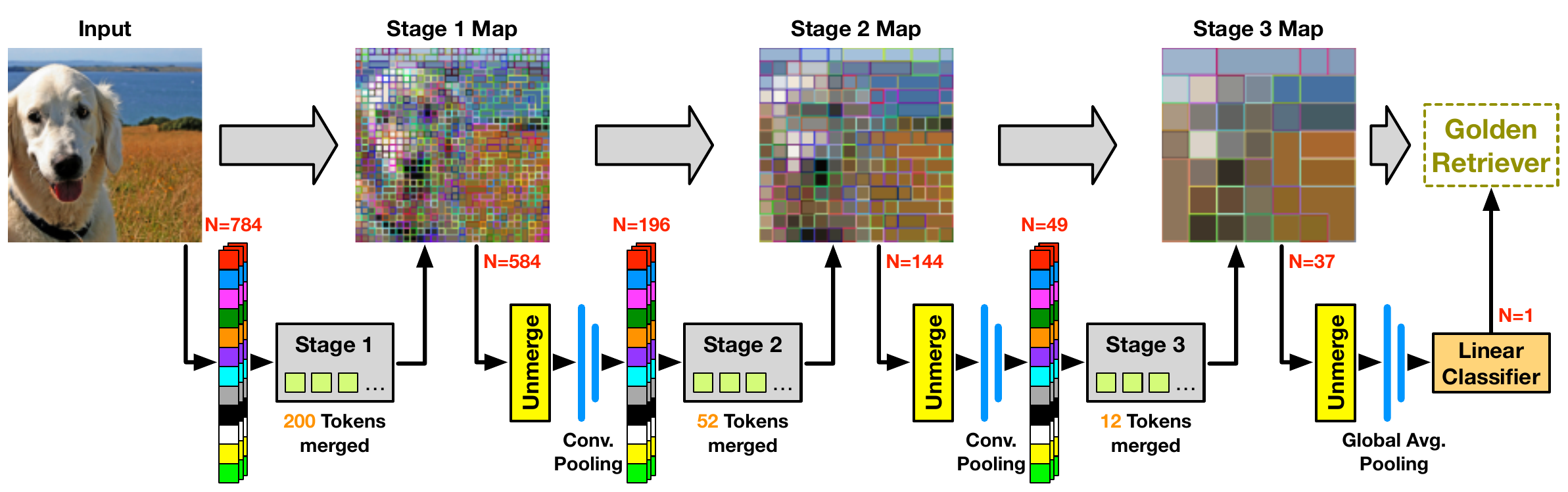}
    \caption{An overview of the proposed \MergeOver approach, integrating the recursive weight-sharing of SReT with the token merging of ToMe. The initial token sequence of length $N = 784$ is reduced using a stage-wise single-shot schedule. Crucially, the spatial layout is reconstructed prior to each inter-stage convolutional pooling layer through the \emph{Unmerge} block.}
    \label{fig:integration}
\end{figure*}

\Cref{fig:integration} illustrates the complete technical integration of SReT and ToMe, detailing how tokens are reduced throughout the complete forward pass. The inner workings of different steps are described.

\subsection{Spatial layout constraints}
\begin{figure}[htbp]
    \centering
    \includegraphics[width=0.9\linewidth]{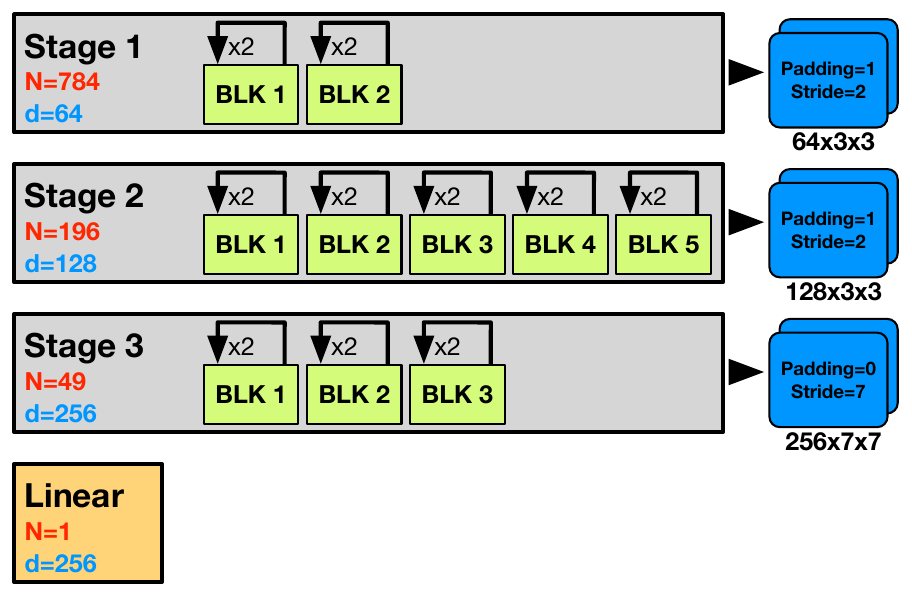}
    \caption{The hierarchical spatial pyramid architecture of SReT, covering the 3 stages and the Linear classifier stage from \Cref{fig:integration}. Each stage involves Transformer blocks (in orange) with $\times 2$ recursive iterations, leading to convolutional pooling layers (in blue). Token sequence length ($N$) and channel depth ($d$) values are provided throughout. Dynamic token reduction must satisfy the strict spatial downsampling of convolutional pooling layers.}
    \label{fig:sret}
\end{figure}

SReT~\cite{sret-shen2022} inherits its structure directly from PiT~\cite{pit-heo2021}, which utilises a spatial pyramid design. Unlike standard ViTs that split an image into non-overlapping patches using a linear projection, SReT applies overlapping convolutional embeddings to generate its initial token sequence. As depicted in \Cref{fig:sret}, these tokens are then processed through three distinct stages. Each stage contains a set of Transformer blocks that are executed recursively. Between these stages, convolutional pooling layers decrease the spatial sequence length while increasing channel depth. After the final stage, global average pooling compresses the tokens for final classification.

This hierarchical design means that after tokens exit the Transformer blocks of a given stage, they must pass through an inter-stage convolutional pooling layer. Since convolutional kernels operate on structured 2D spatial grids, e.g., $14 \times 14$, they expect a sequence length that can be perfectly reshaped into a square tensor. However, standard token reduction eliminates an arbitrary number of tokens, destroying this perfect square property. Consequently, the sequence becomes structurally invalid for the convolutional kernels, triggering a dimensional clash at the end of the stage.

The BSM algorithm of ToMe~\cite{tome-bolya2023} provides a unique mechanism to resolve this clash. During a merge operation, ToMe returns both a merging function and a corresponding \emph{Unmerge} function. Crucially, this \emph{Unmerge} function does not reverse the aggregation of the features. Instead, it restores the spatial dimensions of the sequence by duplicating the merged features back into their original spatial coordinates. This implies that regardless of how aggressively a sequence is compressed, the initial sequence length can be restored, provided that the \emph{Unmerge} functions are stored. This property is highly beneficial for SReT, offering a direct solution to satisfy the geometric requirements of inter-stage convolutional pooling.

\begin{figure}[htbp]
    \centering
    \includegraphics[width=0.9\linewidth]{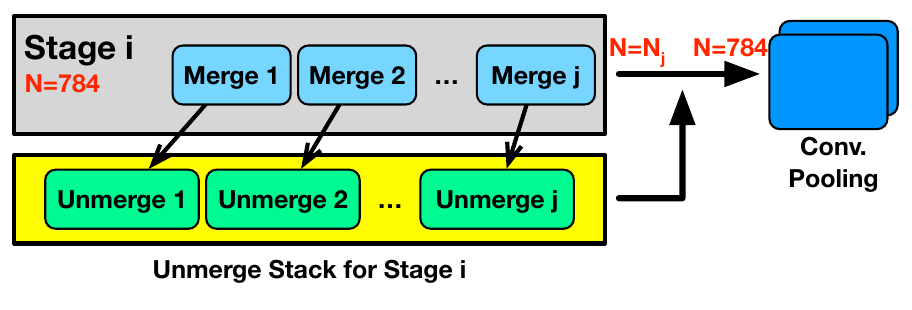}
    \caption{The \emph{Unmerge stack} mechanism for token sequence length restoration. By applying \emph{Unmerge} operations, the original length ($N = 784$) is restored to satisfy the spatial grid requirements of the convolutional pooling layer.}
    \label{fig:unmerge}
\end{figure}

To utilise this capability, we implement an \emph{Unmerge stack} as shown in \Cref{fig:unmerge}. As the token reduction schedule dynamically executes merge operations within a stage, the corresponding \emph{Unmerge} operations are pushed onto a last-in-first-out stack. Before the compressed sequence reaches the inter-stage convolutional pooling layer, the tensor is built back up to its original dimensions by applying the stored \emph{Unmerge} operations in strict reverse order.

This mechanism allows the attention and Multilayer Perceptron (MLP) blocks to reap the computational benefits of processing a compressed token sequence, while ensuring that the convolutional pooling layers receive grid-aligned tokens. The mechanism introduces operational overhead associated with tracking the \emph{Unmerge} functions and restoring the token sequence at the end of each stage.

\subsection{Merging constraints}
\begin{figure}[htbp]
    \centering
    \includegraphics[width=0.9\linewidth]{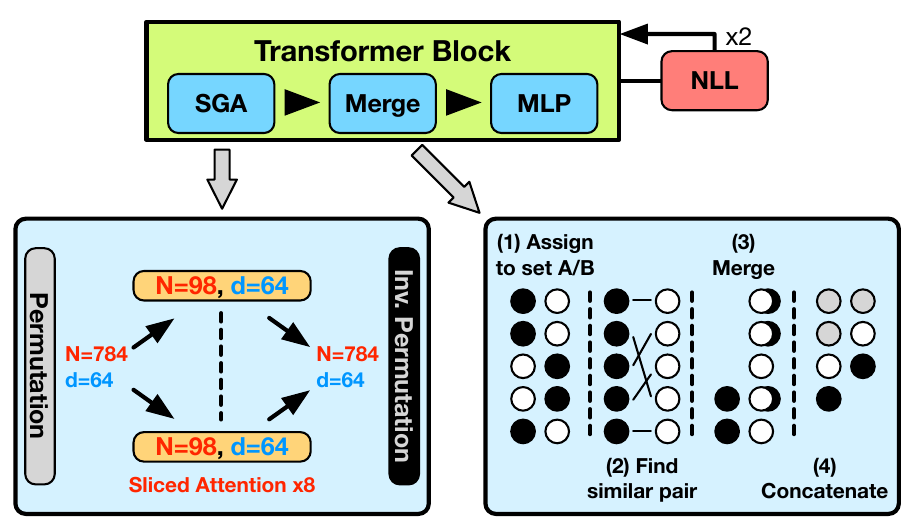}
    \caption{The \MergeOver Transformer block (SReT+ToMe), with the token merging module inserted between the Sliced Group Attention (SGA) and the Multilayer Perceptron (MLP) layers. Detailed expansions illustrate the spatial permutations and grouped attention within SGA, alongside the bipartite matching steps used to merge tokens. The Non-Linear projection Layer (NLL) is part of the recursion.}
    \label{fig:sret-tome-block}
\end{figure}

With the geometric requirements of the inter-stage convolutional pooling layers resolved by the \emph{Unmerge stack}, we must now address the internal constraints of the Transformer blocks. Due to the SGA mechanism of SReT and the BSM algorithm of ToMe, the number of tokens merged at every iteration ($r$) cannot be set arbitrarily. As depicted in \Cref{fig:sret-tome-block}, the token merging operation is inserted between the SGA and the MLP layers. Consequently, the reduction rate $r$ must satisfy both the group divisibility constraints of the subsequent SGA operation and the BSM's bipartite matching limits.

Within the SGA, the token sequence is divided equally into groups. As illustrated in \Cref{fig:sret-tome-block}, attention is computed separately for each group. Thus, a strict constraint is that the sequence length entering the block must be evenly divisible by the designated \emph{group number} ($g$) for that iteration, and there must be at least one token per group. For example, an incoming sequence length of $N = 784$ complies perfectly with a group number of $g = 8$, as each group receives exactly 98 tokens (\Cref{fig:sret-tome-block}). Conversely, sequence lengths of $N = 15$ or $N = 7$ would trigger immediate tensor shape mismatches.

Since SReT blocks are executed recursively, the compressed sequence length exiting the current iteration ($N - r$) is the input sequence for the next iteration. Furthermore, SReT defines a different, generally lower group number ($g_2$) for such next iterations to allow each group to access more tokens, expanding the receptive field of the attention mechanism. To ensure the compressed sequence satisfies the requirements of SGA across all recursive passes, $N - r$ must be divisible by the Least Common Multiple (LCM) of the group numbers ($g_1, g_2, ...$). Moreover, ToMe executes bipartite matching by first dividing the tokens into two sets (\Cref{fig:sret-tome-block}). Since a token from one set is merged into a target token from the other, the algorithm cannot merge more than half of the total available tokens in a single iteration. Therefore, $r$ must also be upper bounded by $\lfloor N/2 \rfloor$, resulting in the merging constraints, which can be formalised as:
\begin{equation}
    (N - r) \bmod g = 0, \;\; N - r \ge g, \;\; g = \operatorname{LCM}(g_1, g_2), \;\; r \le \lfloor N/2 \rfloor.
    \label{eq:r-constraint-final}
\end{equation}

\subsection{Spatial tracking}
As tokens are merged across the recursive stages of SReT, a single token comes to represent an increasingly large number of original patches. While standard ViT and SReT architectures treat all tokens equally, ToMe introduces a parallel token-size vector ($s$) to track the accumulated \emph{mass} of every token to ensure proportional feature aggregation. Proportional attention (\Cref{eq:proportional-attention}) is achieved by scaling the Key matrix using $s$, prior to the Softmax operation. To integrate this into SReT, we initialise a \emph{mass} tensor of ones at the beginning of each forward pass. This tensor is updated across all layers and merge operations, to track the exact weight of each token.

During the forward pass, redundant tokens are identified via BSM. Crucially, we evaluate token similarity using the unweighted Key matrix ($K$) extracted from the attention layer. Isolating the similarity metric from the token mass ensures that merges are determined purely by semantic feature similarity, preventing \enquote{heavy} tokens with high accumulated mass from dominating the matching process.

Once BSM determines the optimal pairs, we merge the tokens using a weighted average. Since the tokens accumulate varying \enquote{masses} (the number of original spatial patches they represent), applying a naive, unweighted average would create a representational imbalance. Specifically, it would allow single-patch tokens equal influence over the merged result as multi-patch tokens. To prevent this, we pre-weight the input feature tensor by its corresponding mass via element-wise multiplication. The merging function then applies a sum reduction to independently aggregate both the weighted features and the parallel mass metrics. Finally, dividing the aggregated feature tensor by the updated mass tensor yields the correct weighted average representation.

A final structural challenge arises from SReT's use of token permutations. As depicted in \Cref{fig:sret-tome-block}, SGA applies permutations and inverse permutations to the input sequence to shift the attention window across recursive passes, ensuring global token interaction. If the mass tensor is not permuted identically, the weights will misalign with their corresponding tokens, corrupting the proportional attention. By decoupling the token-mass tensor ($s$) and routing it in parallel through the exact same permutation and inverse permutation functions as the feature sequence, we guarantee that the mass metrics remain synchronised with their corresponding tokens.

\subsection{Token reduction scheduling}
The original ToMe framework utilises two token reduction schedules: \emph{constant} and \emph{linearly decreasing}~\cite{tome-bolya2023}. These schedules are defined using an absolute token count over a global network-depth index. The constant schedule merges a fixed number of tokens ($r$) at every merge-enabled block: 
\begin{equation}
    r_d^{\mathrm{const}} = r,
    \label{eq:constant}
\end{equation}
where $r_d^{\mathrm{const}}$ is the number of tokens removed at block depth $d$. The linearly decreasing schedule starts by merging $2r$ tokens in the first layer and reduces this amount to $0$ by the final layer:
\begin{equation}
    r_d^{\mathrm{lin}} =
    \left\lfloor
    2r\left(1 - \frac{d-1}{D-1}\right)
    \right\rfloor,
    \qquad
    d = 1, \ldots, D,
    \label{eq:linear}
\end{equation}
where $D$ is the total number of blocks over which token merging is applied. Thus, $r_1^{\mathrm{lin}} = 2r$ and $r_D^{\mathrm{lin}} = 0$. While both schedules remove nearly the same total number of tokens for a given value of $r$, they distribute it differently over network depth. We retain these schedules as canonical ToMe baselines.

To account for the changing sequence lengths of the SReT pyramid, we additionally formulate token reduction schedules relative to the initial sequence length of each stage: \emph{stage-wise exponential} and \emph{stage-wise single-shot}. Let $s$ denote the stage, $N_s$ its initial sequence length, $D_s$ its number of blocks, and $d=1,\ldots,D_s$ the local block-depth index.

The stage-wise exponential schedule requests progressively smaller reductions:
\begin{equation}
    r_{s,d}^{\mathrm{exp}} =
    \left\lfloor 
       \rho_{\mathrm{exp}} \cdot N_s \cdot \alpha^{d-1}
    \right\rfloor,
    \qquad
    0 < \alpha < 1,
    \label{eq:exponential}
\end{equation}
where $\rho_{\mathrm{exp}}$ is the stage-initial reduction fraction and $\alpha$ is the decay factor. Larger values of $\alpha$ produce slower decay, whereas smaller values concentrate the requested reduction near the beginning of each stage. 

Conversely, a stage-wise single-shot schedule performs token reduction only at the first block of each stage:
\begin{equation}
    r_{s,d}^{\mathrm{shot}} =
    \begin{cases}
        \left\lfloor \rho_{\mathrm{shot}} \cdot N_s \right\rfloor,
            & d=1, \\[3pt]
        0,   & d>1,
    \end{cases}
    \label{eq:single-shot}
\end{equation}
where $\rho_{\mathrm{shot}}$ is the stage-initial reduction fraction. This schedule maintains a fixed sequence length throughout the remaining blocks of the stage. Both schedules reset at stage boundaries.

\begin{figure}[htbp]
    \centering
    \includegraphics[width=\linewidth]{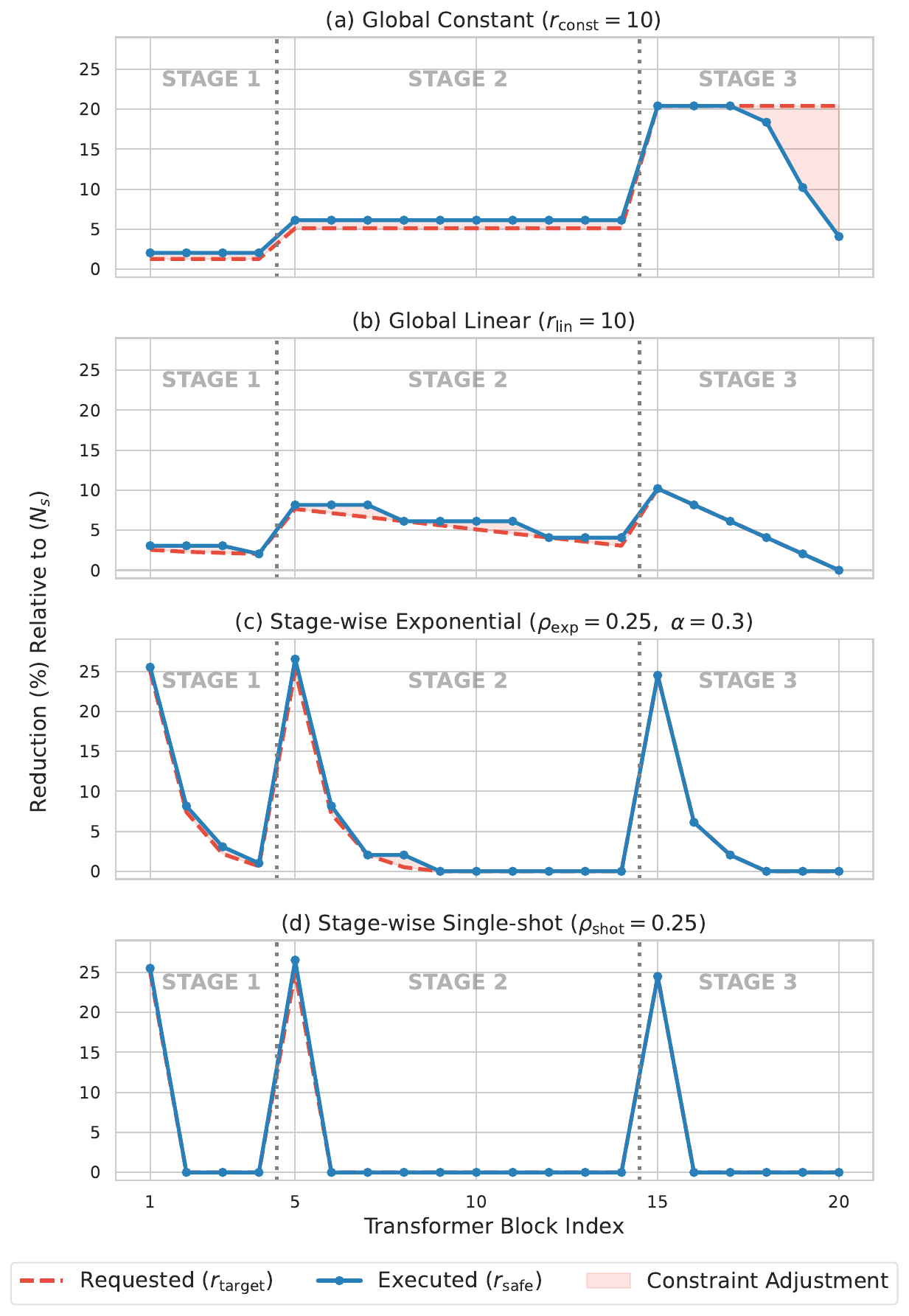}
    \caption{Normalised token reduction trajectories per schedule. Relative to the stage-initial sequence length ($N_\mathrm{s}$), reductions increase at later stages for global schedules, while stage-wise schedules show near-identical trajectories throughout all stages. The constraint adjustments from $r_\mathrm{target}$ to $r_\mathrm{safe}$ illustrate the merging constraints of Equation~\ref{eq:r-constraint-final}.}
    \label{fig:reduction-combined}
\end{figure}

Since the canonical and stage-wise schedules differ in both parameterisation and depth indexing, comparisons between them only evaluate the scheduling trajectories. \Cref{fig:reduction-combined} illustrates these differences in trajectory for the four schedules. For the global schedules (constant and linear), the reduction percentages relative to the initial starting sequence length of each stage ($N_\mathrm{s}$) increase after each stage even though the absolute number of tokens removed remains identical (constant) or decreases (linear). Conversely, the stage-wise schedules show near identical trajectories for all stages. For every schedule, the requested reduction $r_{\mathrm{target}}$ is adjusted to the nearest feasible reduction $r_{\mathrm{safe}}$ satisfying the group-divisibility and bipartite-matching constraints of \Cref{eq:r-constraint-final}.

\section{Experimental setup}
\label{sec:experimental-setup}
We detail the experimental setup used to evaluate our proposed approach.

\subsection{Dataset}
All experiments are evaluated on the standard ImageNet-1K validation dataset (ILSVRC 2012)~\cite{imagenet-deng2009}, which contains 50\,000 images distributed evenly across 1\,000 object classes. We preprocess all images using the standard ViT pipeline: resizing images to a 256-pixel shorter edge, extracting a $224 \times 224$ centre crop, and normalising the pixel values using the standard channel-wise ImageNet mean and standard deviation.

\subsection{Hardware and software environment}
Evaluations are conducted using an NVIDIA GeForce RTX 4060 Ti GPU, an Intel Core Ultra 9 285K CPU (x86) with 64 GB of DDR-5 RAM, and a separate Raspberry Pi 5 (ARM) with 16 GB of RAM to demonstrate embedded deployment. The software stack consists of Python~3.10, CUDA~13.0, PyTorch~2.11.0, Torchvision~0.26.0, and \texttt{timm}~0.4.12. Theoretical complexity, measured in Floating-Point Operations (FLOPs), is computed using \texttt{thop}~0.1.1. Hardware throughput is measured using native CUDA timing events for GPU evaluations and custom performance snapshot utilities for CPU evaluations.

Specific execution parameters are enforced for accurate and comparative benchmarking. For GPU evaluations, cuDNN auto-tuning is enabled to optimise backend execution kernels. For CPU evaluations, strict GPU isolation is enforced, and computation is explicitly constrained to four hardware threads. Moreover, MKLDNN is enabled for CPU environments.

\subsection{Metrics}
To evaluate the trade-offs introduced by the proposed integration, we measure hardware efficiency (latency (ms), throughput (img/s), and PAM (MB)), theoretical complexity (FLOPs (G) and parameter counts (M)), as well as task performance (Top-1 accuracy (\%)).

We evaluate GPU throughput and PAM for Batch Sizes (BS) ranging from 1 to 128 in powers of 2. We simulate single-stream edge inference ($BS = 1$) and assess peak parallelisation ($BS = 128$), while evaluating the variations in performance for the intermediate batch sizes. For experiments involving a CPU, latency, throughput, and memory are evaluated for batch sizes ranging from 1 to 16 in powers of 2, to reflect edge-computing relevance.

GPU PAM is measured using PyTorch's native CUDA allocator. CPU memory is reported separately as the change in process Resident Set Size ($\Delta$RSS) observed during inference, relative to the pre-inference process state. Unlike GPU PAM, $\Delta$RSS is a process-level measurement that includes tensor allocations, allocator caching, backend workspaces, and other resident memory. It therefore provides an estimate of changes in CPU process-memory usage rather than an isolated measurement of activation memory. Thus, the GPU and CPU memory values are not directly comparable and are only provided for reference.

We measure per-batch inference latency after a warm-up phase (20 iterations for GPU, 5 iterations for CPU), followed by evaluation iterations (100 iterations for GPU, 50 iterations for CPU). The median latency across the evaluation iterations is reported. Throughput is derived from the median batch latency, and thus latency and throughput are not independent measurements.

\subsection{Baselines}
We compare our results against the uncompressed SReT-Tiny-Distill~\cite{sret-shen2022}, serving as a recursive hierarchical ViT baseline. Pre-trained weights for SReT are acquired directly from the authors' public repository~\cite{sret-repo}.

\subsection{Token reduction scheduling strategies}
To evaluate the impact of different token reduction scheduling strategies, we benchmark all four schedule types: \emph{constant}, \emph{linear}, \emph{exponential}, and \emph{single-shot}. We intentionally select the following configurations to expose the performance boundaries of recursive architectures:
\begin{itemize}
    \item \textbf{Constant Schedule ($r_{\mathrm{const}}$):} We evaluate the canonical ToMe global constant schedule at $r_{\mathrm{const}}\in\{10,20\}$, representing \emph{light} and \emph{heavy} reductions.
    \item \textbf{Linear Schedule ($r_{\mathrm{lin}}$):} We evaluate the canonical ToMe global linear schedule at $r_{\mathrm{lin}}\in\{10,20\}$. These matched configurations test whether aggressively removing tokens early benefits hierarchical, recursive architectures compared to their constant counterparts.
    \item \textbf{Exponential Schedule ($\rho_{\mathrm{exp}}, \alpha$):} We evaluate a \emph{moderate reduction} ($\rho_{\mathrm{exp}} = 0.25, \alpha = 0.3$) and an \emph{aggressive reduction} ($\rho_{\mathrm{exp}} = 0.40, \alpha = 0.3$). We fix $\alpha$ to isolate the effect of the stage-initial reduction fraction while maintaining a common exponential trajectory. 
    \item \textbf{Single-shot Schedule ($\rho_{\mathrm{shot}}$):} We evaluate $\rho_{\mathrm{shot}}\in\{0.25,0.40\}$, matching the initial reduction fractions of the exponential configurations. This enables a direct comparison between the continued exponential reductions and stage-initial reductions followed by fixed sequence lengths.
\end{itemize}

\section{Benchmarking results and discussion}
\label{sec:results}
Independent of the chosen processing platform (GPU, x86 CPU, ARM CPU) or batch size, \Cref{tab:batch-independent-metrics} details the evaluation results for baseline model performance metrics. \Cref{tab:gpu-batch-dependent-results}, \Cref{tab:cpu-batch-dependent-results}, and \Cref{tab:pi-batch-dependent-results} in \Cref{app:extended-results} detail the performance metrics for all platforms and batch sizes.
\begin{table}[htbp]
\centering
\caption{Platform and batch size independent model performance results on ImageNet-1K. Our selected configuration is highlighted. Percentage changes are relative to the SReT baseline.}
\label{tab:batch-independent-metrics}
\resizebox{\columnwidth}{!}{%
    \begin{tabular}{llll}
        \toprule
        \textbf{Configuration} & 
        \textbf{Top-1 Acc. (\%)} & 
        \textbf{Params. (M)} & 
        \textbf{FLOPs (G)} \\
        \midrule
        \rowHLBaseline
        \textit{SReT baseline}                   & 77.39                  & 4.76   & 1.91 \\
        $r_{\mathrm{const}}=10$                  & 70.95\negNum{6.44}     & \ditto & 1.32\negG{30.8} \\
        $r_{\mathrm{const}}=20$                  & 40.97\negNum{36.42}    & \ditto & 1.06\negG{44.5} \\
        $r_{\mathrm{lin}}=10$                    & 74.74\negNum{2.65}     & \ditto & 1.46\negG{23.5} \\
        $r_{\mathrm{lin}}=20$                    & 14.76\negNum{62.63}    & \ditto & 1.07\negG{43.9} \\
        $\rho_{\mathrm{exp}}=0.25,\ \alpha=0.3$  & 74.58\negNum{2.81}     & \ditto & 1.35\negG{29.3} \\
        $\rho_{\mathrm{exp}}=0.40,\ \alpha=0.3$  & 66.28\negNum{11.11}    & \ditto & 1.04\negG{45.4} \\
        \rowHLBest
        $\rho_{\mathrm{shot}}=0.25$              & 75.92\negNum{1.47}     & \ditto & 1.49\negG{22.0} \\
        $\rho_{\mathrm{shot}}=0.40$              & 72.36\negNum{5.03}     & \ditto & 1.25\negG{34.3} \\
        \bottomrule
    \end{tabular}%
}
\end{table}

\subsection{Token reduction schedules on GPU}
\begin{figure}[htbp]
    \centering
    \includegraphics[width=\linewidth]{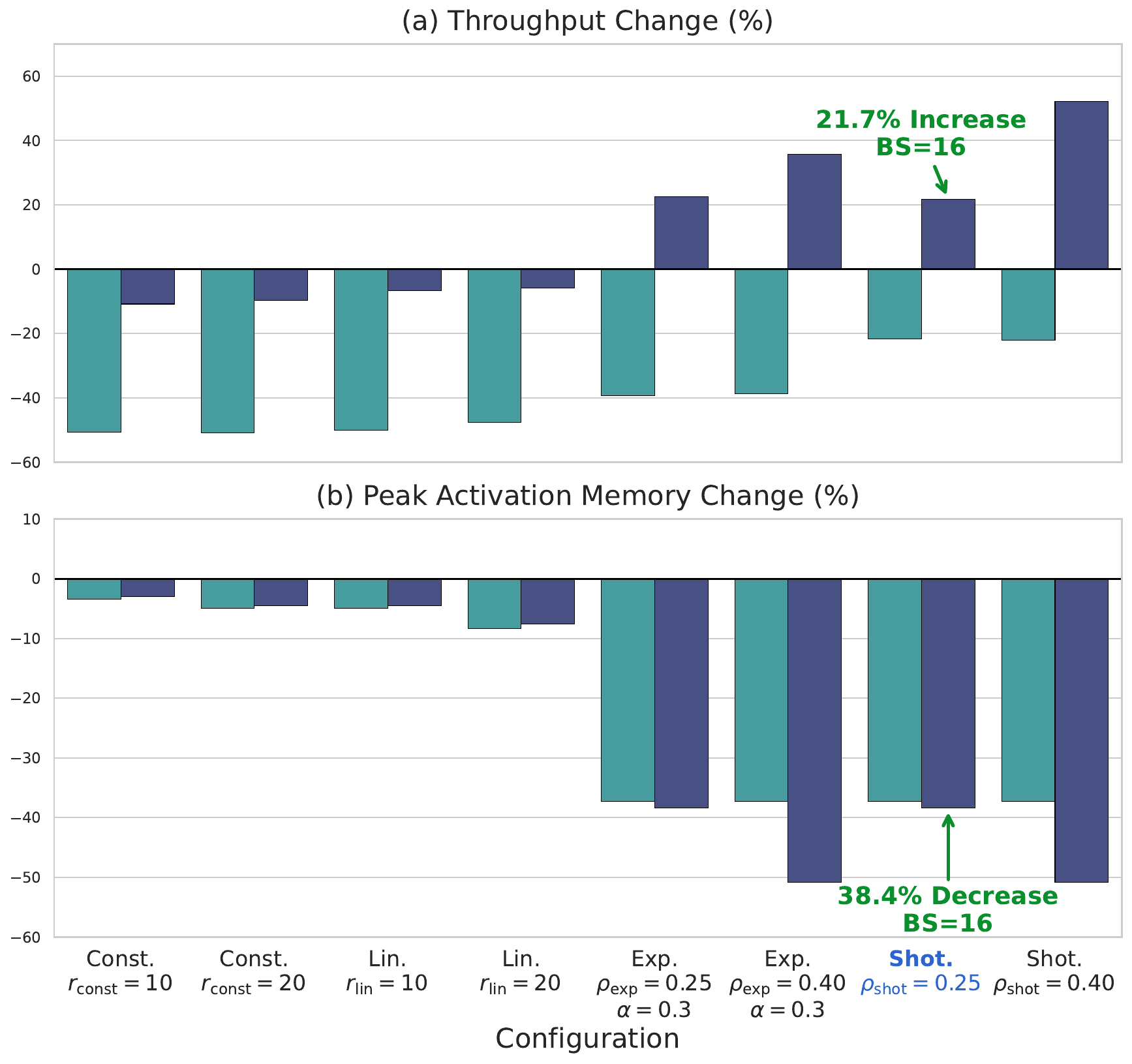}
    \caption{GPU performance results compared to the SReT baseline. Batch sizes 1 and 16 are represented with light and dark shades, respectively. Our chosen configuration shows a 21.7\% increase in throughput and a 38.4\% decrease in PAM for $BS = 16$.}
    \label{fig:gpu-improvements-bar}
\end{figure}

\Cref{tab:gpu-selected-batch-results} and \Cref{fig:gpu-improvements-bar} summarise the throughput and PAM results for all tested configurations for batch sizes 1 and 16. At batch size 16, the global reduction schedules (constant and linear) do not improve throughput and show minimal reductions in PAM, while the stage-wise reduction schedules (exponential and single-shot) show meaningful improvements. Since the global and stage-wise schedules differ in both parameterisation and depth indexing, we do not attribute these outcomes to schedule shape alone. As illustrated in \Cref{fig:reduction-combined}, the stage-wise schedules express reductions relative to each stage's initial sequence length and reset at stage boundaries, whereas the global schedules operate using absolute token counts over global network depth. Nonetheless, it is observed that between the global schedules, the linear schedule slightly outperforms the constant schedule.
\begin{table}[htbp]
\centering
\setlength{\tabcolsep}{4pt}
\caption{GPU performance on ImageNet-1K at batch sizes 1 and 16. Throughput (ThP) and Peak Activation Memory (PAM) are measured on an NVIDIA GeForce RTX 4060 Ti. Our selected configuration is highlighted. Percentage changes are relative to the SReT baseline.}
\label{tab:gpu-selected-batch-results}
\resizebox{\columnwidth}{!}{%
    \begin{tabular}{lllll}
        \toprule
        & 
        \multicolumn{2}{c}{\textbf{Batch size 1}} & 
        \multicolumn{2}{c}{\textbf{Batch size 16}} \\
        \cmidrule(lr){2-3}
        \cmidrule(lr){4-5}
        \textbf{Configuration} & 
        \textbf{ThP (img/s)} & 
        \textbf{PAM (MB)} & 
        \textbf{ThP (img/s)} & 
        \textbf{PAM (MB)} \\
        \midrule
        \rowHLBaseline
        \textit{SReT baseline}                   & 223.83               & 6.22              & 1304.58               & 99.47 \\
        $r_{\mathrm{const}}=10$                  & 110.24\negR{50.7}    & 6.01\negG{3.3}    & 1163.09\negR{10.8}    & 96.50\negG{3.0} \\
        $r_{\mathrm{const}}=20$                  & 109.88\negR{50.9}    & 5.91\negG{4.9}    & 1177.86\negR{9.7}     & 94.95\negG{4.5} \\
        $r_{\mathrm{lin}}=10$                    & 111.56\negR{50.2}    & 5.91\negG{4.9}    & 1218.04\negR{6.6}     & 94.95\negG{4.5} \\
        $r_{\mathrm{lin}}=20$                    & 117.12\negR{47.7}    & 5.70\negG{8.3}    & 1227.17\negR{5.9}     & 91.89\negG{7.6} \\
        $\rho_{\mathrm{exp}}=0.25,\ \alpha=0.3$  & 135.80\negR{39.3}    & 3.90\negG{37.3}   & 1597.22\posG{22.4}    & 61.27\negG{38.4} \\
        $\rho_{\mathrm{exp}}=0.40,\ \alpha=0.3$  & 136.99\negR{38.8}    & 3.90\negG{37.3}   & 1770.14\posG{35.7}    & 48.94\negG{50.8} \\
        \rowHLBest
        $\rho_{\mathrm{shot}}=0.25$              & 175.33\negR{21.7}    & 3.90\negG{37.3}   & 1588.00\posG{21.7}    & 61.27\negG{38.4} \\
        $\rho_{\mathrm{shot}}=0.40$              & 174.42\negR{22.1}    & 3.90\negG{37.3}   & 1985.57\posG{52.2}    & 48.94\negG{50.8} \\
        \bottomrule
    \end{tabular}%
}
\end{table}

At batch size 1, however, \MergeOver degrades GPU throughput for every tested configuration. This behaviour is consistent with the GPU requiring sufficient parallel work to amortise the added overhead of \MergeOver. At higher batch sizes, this overhead can be distributed across a larger amount of parallel computation, while, during single-stream inference, this overhead is more prominent. This batch-dependent crossover is observed across the batch sizes from 1 to 128, with the detailed results reported in \Cref{tab:gpu-batch-dependent-results} in \Cref{app:extended-results}. Low-level profiling and kernel optimisation are therefore warranted as future work. 

Furthermore, \Cref{tab:batch-independent-metrics} shows that a configuration with fewer FLOPs does not necessarily achieve higher throughput. The $\rho_\mathrm{exp} = 0.40, \alpha = 0.3$ configuration has an estimated cost of 1.04 GFLOPs and a throughput increase of 35.7\% at a batch size of 16. In contrast, the $\rho_\mathrm{shot} = 0.40$ configuration shows a higher throughput increase of 52.2\% despite having a higher estimated cost of 1.25 GFLOPs. The FLOPs estimate does not adequately represent costs such as BSM matching, token-mass tracking, and \emph{Unmerge} operations. Moreover, the single-shot schedule invokes token merging in fewer blocks than the exponential schedule. We therefore treat FLOPs as an approximate measure of tensor arithmetic rather than a predictor of hardware performance.

When evaluating the performance of the schedules, \emph{we adopt a maximum top-1 accuracy loss of 2 percentage points as an accuracy-preserving criterion}. Under this criterion, we select the stage-wise single-shot configuration $\rho_{\mathrm{shot}} = 0.25$, which reduces top-1 accuracy by 1.47 percentage points. However, this does not imply that the configuration is globally optimal. More aggressive configurations result in greater hardware performance at the cost of larger accuracy drops. The selected configuration reduces GPU throughput by 21.7\% at batch size 1, while reducing PAM by 37.3\%. At batch size 16, it increases throughput by 21.7\% and reduces PAM by 38.4\%. We also note that, in the evaluated configurations for the GPU, PAM is determined primarily by the initial reduction fraction ($\rho_{\mathrm{exp}}$ or $\rho_{\mathrm{shot}}$), as the largest activation occurs near the beginning of each stage and subsequent reductions operate on already shortened sequences. This explains why exponential and single-shot schedules with the same initial reduction fraction produce similar PAM. \MergeOver therefore provides a favourable efficiency trade-off under batched execution, but does not accelerate single-stream GPU inference.

\subsection{Token reduction schedules on x86 CPU}
\begin{figure}[htbp]
    \centering
    \includegraphics[width=\linewidth]{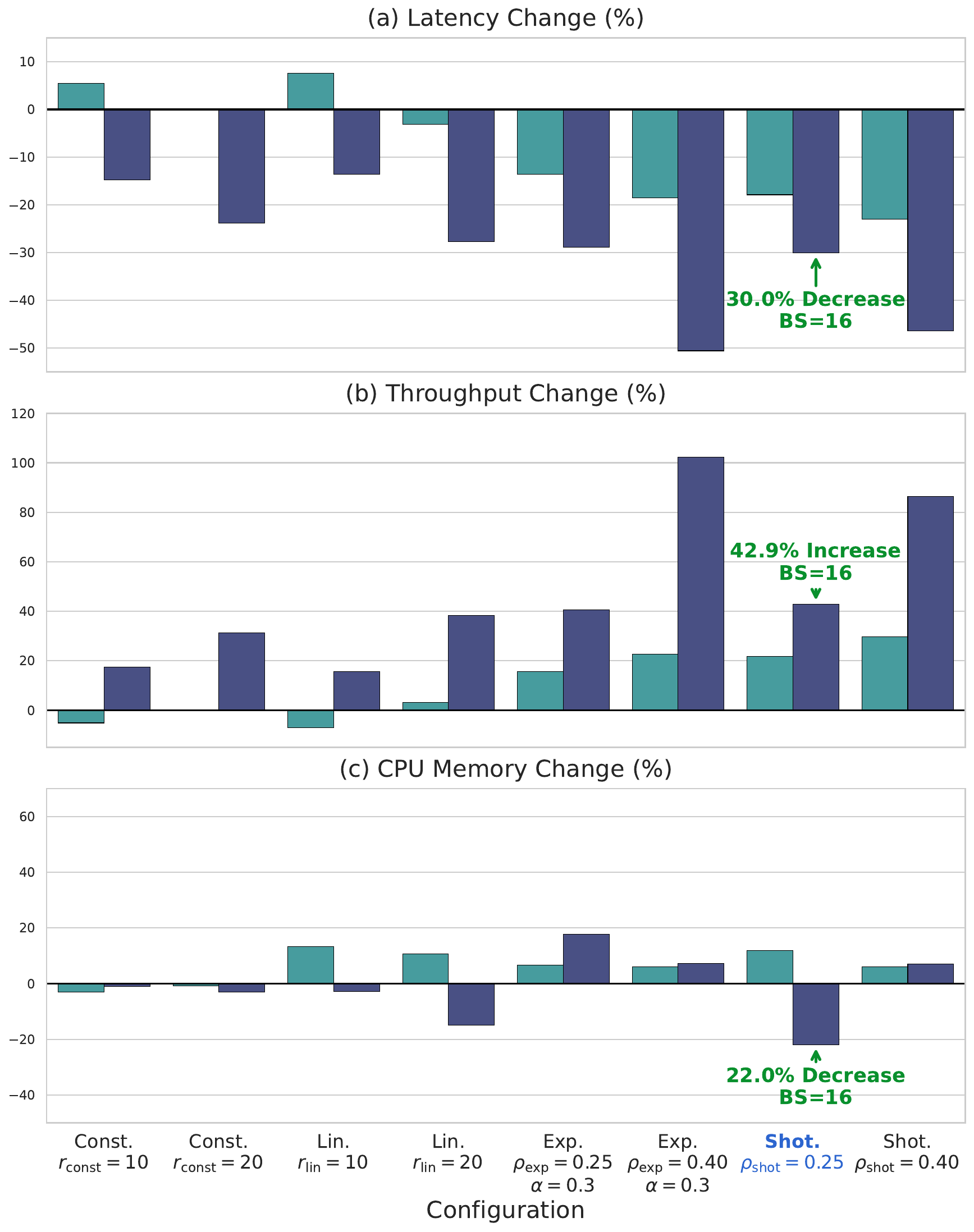}
    \caption{x86 CPU performance results compared to the SReT baseline. Batch sizes 1 and 16 are represented with light and dark shades, respectively. Our chosen configuration shows a 30.0\% decrease in latency, a 42.9\% increase in throughput, and a 22.0\% decrease in process-memory change ($\Delta$RSS) for $BS = 16$.}
    \label{fig:cpu-improvements-bar}
\end{figure}

\Cref{tab:x86-selected-batch-results} and \Cref{fig:cpu-improvements-bar} summarise the latency and process-memory results on the x86 CPU. In contrast to the GPU results, the stage-wise schedules improve latency at both batch sizes 1 and 16. At batch size 1, the constant and linear schedules with $r = 10$ increase latency by 5.4\% and 7.6\%, respectively, while their more aggressive variants provide reductions of only 0.1\% and 3.1\%. By comparison, the stage-wise schedules reduce latency by 13.6--23.0\%. This indicates that the single-stream overhead observed on the GPU does not transfer directly to CPU execution and that the efficiency of \MergeOver is platform-dependent.
\begin{table}[htbp]
\centering
\setlength{\tabcolsep}{4pt}
\caption{x86 CPU performance on ImageNet-1K at batch sizes 1 and 16. Latency (Lat) and process-memory change ($\Delta$RSS) are measured on an Intel Core Ultra 9 285K using four threads. Our selected configuration is highlighted. Percentage changes are relative to the SReT baseline.}
\label{tab:x86-selected-batch-results}
\resizebox{\columnwidth}{!}{%
    \begin{tabular}{lllll}
        \toprule
        &
        \multicolumn{2}{c}{\textbf{Batch size 1}} &
        \multicolumn{2}{c}{\textbf{Batch size 16}} \\
        \cmidrule(lr){2-3}
        \cmidrule(lr){4-5}
        \textbf{Configuration} &
        \textbf{Lat (ms)} &
        \textbf{$\Delta$RSS (MB)} &
        \textbf{Lat (ms)} &
        \textbf{$\Delta$RSS (MB)} \\
        \midrule
        \rowHLBaseline
        \textit{SReT baseline}                      & 14.94             & 19.94             & 144.32                & 92.97 \\
        $r_{\mathrm{const}}=10$                     & 15.75\posR{5.4}   & 19.35\negG{3.0}   & 122.95\negG{14.8}     & 91.97\negG{1.1} \\
        $r_{\mathrm{const}}=20$                     & 14.92\negG{0.1}   & 19.77\negG{0.9}   & 109.90\negG{23.8}     & 90.08\negG{3.1} \\
        $r_{\mathrm{lin}}=10$                       & 16.08\posR{7.6}   & 22.60\posR{13.3}  & 124.76\negG{13.6}     & 90.41\negG{2.8} \\
        $r_{\mathrm{lin}}=20$                       & 14.47\negG{3.1}   & 22.07\posR{10.7}  & 104.43\negG{27.6}     & 79.02\negG{15.0} \\
        $\rho_{\mathrm{exp}}=0.25,\ \alpha=0.3$     & 12.91\negG{13.6}  & 21.28\posR{6.7}   & 102.65\negG{28.9}     & 109.37\posR{17.6} \\
        $\rho_{\mathrm{exp}}=0.40,\ \alpha=0.3$     & 12.18\negG{18.5}  & 21.13\posR{6.0}   & 71.34\negG{50.6}      & 99.77\posR{7.3} \\
        \rowHLBest
        $\rho_{\mathrm{shot}}=0.25$                 & 12.27\negG{17.9}  & 22.32\posR{11.9}  & 100.98\negG{30.0}     & 72.52\negG{22.0} \\
        $\rho_{\mathrm{shot}}=0.40$                 & 11.51\negG{23.0}  & 21.13\posR{6.0}   & 77.43\negG{46.3}      & 99.52\posR{7.0} \\
        \bottomrule
    \end{tabular}%
}
\end{table}

At batch size 16, all evaluated schedules reduce latency. The global schedules achieve reductions of 13.6--27.6\%, whereas the stage-wise schedules achieve reductions of 28.9--50.6\%. However, these hardware results must be considered alongside the accuracy results in \Cref{tab:batch-independent-metrics}. In particular, the larger latency reductions of $r_{\mathrm{const}} = 20$, $r_{\mathrm{lin}} = 20$, and the configurations using a reduction fraction of $0.40$ come with substantial accuracy losses. Consequently, latency reduction alone does not determine the preferred configuration.

Under our accuracy-preserving criterion, the selected stage-wise single-shot configuration $\rho_{\mathrm{shot}} = 0.25$ reduces latency by 17.9\% at batch size 1 and by 30.0\% at batch size 16. Unlike on the GPU, \MergeOver accelerates both single-stream and batched inference on the x86 CPU. The larger improvement at batch size 16 follows the pattern that the relative cost of token matching and sequence manipulation decreases as the amount of inference computation increases. Since we did not perform low-level profiling or optimisations, we treat this explanation as a hypothesis rather than a confirmed cause.

The reported $\Delta$RSS values measure changes in process resident memory rather than peak activation memory. Their behaviour across schedules and batch sizes do not follow a strict pattern. For example, the selected configuration increases $\Delta$RSS by 11.9\% at batch size 1 but reduces it by 22.0\% at batch size 16. Similar inconsistencies occur among the other configurations. Due to the known limitations of RSS, we do not use our CPU memory measurements to rank the schedules or attribute these changes directly to token reduction. 

\subsection{Token reduction schedules on ARM CPU}
\begin{figure}[htbp]
    \centering
    \includegraphics[width=\linewidth]{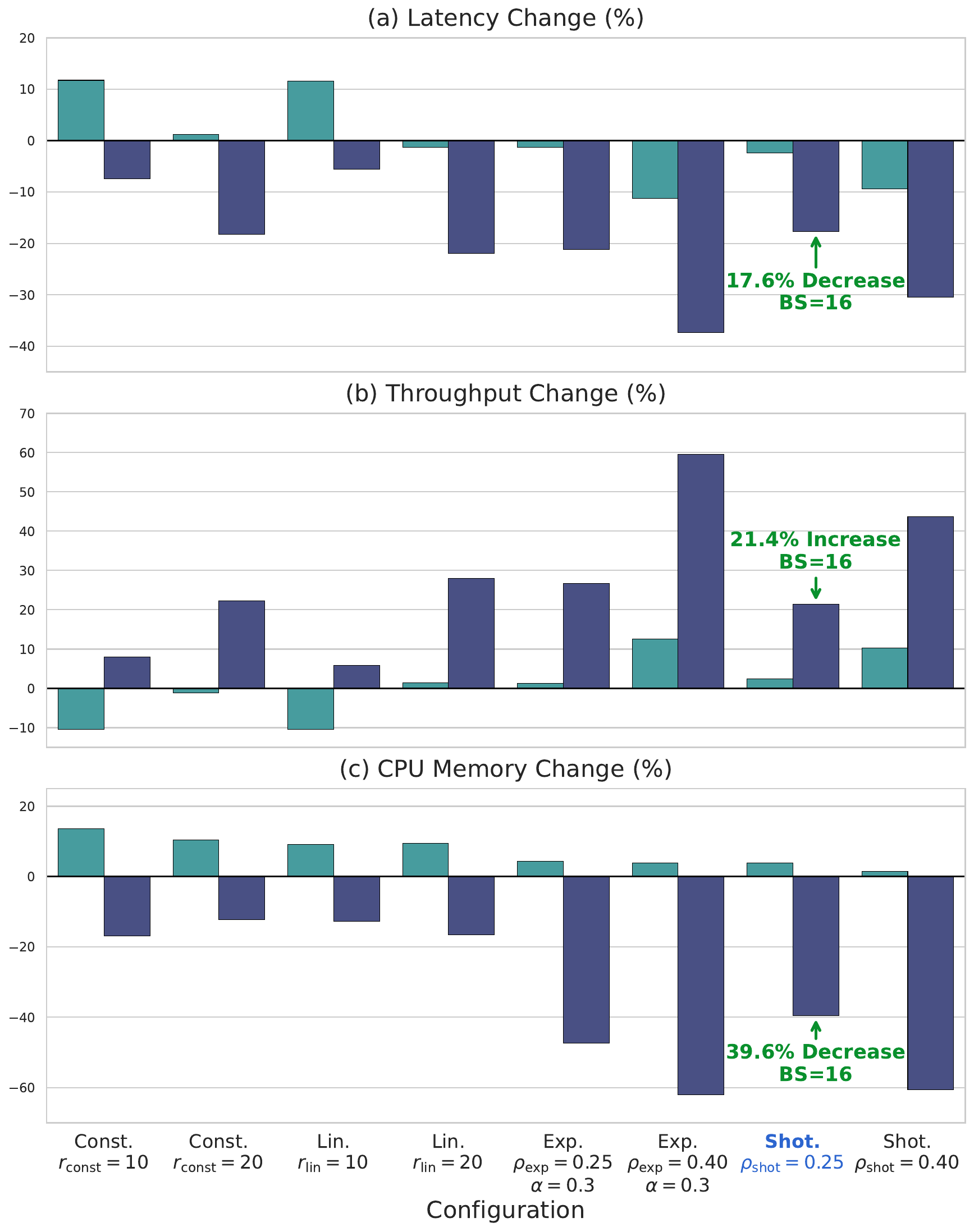}
    \caption{ARM CPU performance results compared to the SReT baseline. Batch sizes 1 and 16 are represented with light and dark shades, respectively. Our chosen configuration shows a 17.6\% decrease in latency, a 21.4\% increase in throughput, and a 39.6\% decrease in process-memory change ($\Delta$RSS) for $BS = 16$.}
    \label{fig:arm-improvements-bar}
\end{figure}

\Cref{tab:arm-selected-batch-results} and \Cref{fig:arm-improvements-bar} summarise the latency and process-memory results on the Raspberry Pi 5. At batch size 1, the light global schedules increase latency by 11.8\% for $r_{\mathrm{const}} = 10$ and 11.6\% for $r_{\mathrm{lin}} = 10$. Their more aggressive counterparts remain close to the baseline, producing a 1.2\% increase and a 1.3\% reduction, respectively. The stage-wise schedules perform more favourably. The exponential and single-shot configurations with an initial reduction fraction of $0.25$ reduce latency by 1.3\% and 2.4\%, respectively. More substantial reductions are obtained with a fraction of $0.40$, but these configurations also come with larger accuracy losses.
\begin{table}[htbp]
\centering
\setlength{\tabcolsep}{4pt}
\caption{ARM CPU performance on ImageNet-1K at batch sizes 1 and 16. Latency (Lat) and process-memory change ($\Delta$RSS) are measured on a Raspberry Pi 5 with 16 GB of RAM. Our selected configuration is highlighted. Percentage changes are relative to the SReT baseline.}
\label{tab:arm-selected-batch-results}
\resizebox{\columnwidth}{!}{%
    \begin{tabular}{lllll}
        \toprule
        &
        \multicolumn{2}{c}{\textbf{Batch size 1}} &
        \multicolumn{2}{c}{\textbf{Batch size 16}} \\
        \cmidrule(lr){2-3}
        \cmidrule(lr){4-5}
        \textbf{Configuration} &
        \textbf{Lat (ms)} &
        \textbf{$\Delta$RSS (MB)} &
        \textbf{Lat (ms)} &
        \textbf{$\Delta$RSS (MB)} \\
        \midrule
        \rowHLBaseline
        \textit{SReT baseline}                      & 142.82            & 21.89             & 1236.71               & 163.23 \\
        $r_{\mathrm{const}}=10$                     & 159.61\posR{11.8} & 24.86\posR{13.6}  & 1145.12\negG{7.4}     & 135.69\negG{16.9} \\
        $r_{\mathrm{const}}=20$                     & 144.50\posR{1.2}  & 24.16\posR{10.4}  & 1011.14\negG{18.2}    & 143.36\negG{12.2} \\
        $r_{\mathrm{lin}}=10$                       & 159.41\posR{11.6} & 23.88\posR{9.1}   & 1167.84\negG{5.6}     & 142.53\negG{12.7} \\
        $r_{\mathrm{lin}}=20$                       & 140.90\negG{1.3}  & 23.97\posR{9.5}   & 965.58\negG{21.9}     & 136.25\negG{16.5} \\
        $\rho_{\mathrm{exp}}=0.25,\ \alpha=0.3$     & 140.94\negG{1.3}  & 22.83\posR{4.3}   & 975.44\negG{21.1}     & 85.98\negG{47.3} \\
        $\rho_{\mathrm{exp}}=0.40,\ \alpha=0.3$     & 126.84\negG{11.2} & 22.73\posR{3.8}   & 774.81\negG{37.3}     & 61.89\negG{62.1} \\
        \rowHLBest
        $\rho_{\mathrm{shot}}=0.25$                 & 139.42\negG{2.4}  & 22.75\posR{3.9}   & 1018.61\negG{17.6}    & 98.66\negG{39.6} \\
        $\rho_{\mathrm{shot}}=0.40$                 & 129.49\negG{9.3}  & 22.23\posR{1.6}   & 860.18\negG{30.4}     & 64.33\negG{60.6} \\
        \bottomrule
    \end{tabular}%
}
\end{table}

At batch size 16, all evaluated schedules reduce latency. The global schedules provide reductions between 5.6\% and 21.9\%, while the stage-wise schedules provide reductions between 17.6\% and 37.3\%. As with the x86 results, the most aggressive configuration does not necessarily provide the most favourable accuracy--latency trade-off. In particular, $\rho_{\mathrm{exp}} = 0.40, \alpha = 0.3$ achieves the largest latency reduction of 37.3\%, but reduces top-1 accuracy by 11.11 percentage points.

Under our accuracy-preserving criterion, the selected stage-wise single-shot configuration $\rho_{\mathrm{shot}} = 0.25$ reduces latency by 2.4\% at batch size 1 and by 17.6\% at batch size 16. The limited improvement at batch size 1 shows that \MergeOver provides only a modest benefit for single-image inference on the Raspberry Pi. Its larger benefit at batch size 16 follows the cross-platform pattern that token reduction becomes more favourable under batched execution, where the computational savings can more effectively offset the additional merging and tracking operations.

Similar to the x86 results, the ARM $\Delta$RSS measurements also depend strongly on batch size. Every evaluated configuration increases $\Delta$RSS at batch size 1, whereas all configurations reduce it at batch size 16. For the selected configuration, $\Delta$RSS increases by 3.9\% at batch size 1 and decreases by 39.6\% at batch size 16.

\section{Limitations and future work}
\label{sec:future-work}
While we successfully demonstrate the viability of post-training dynamic token reduction in recursive architectures, several limitations present clear directions for future work.

First, the evaluation is restricted to SReT-Tiny-Distill. Consequently, we cannot separate effects caused by recursive weight sharing from those caused by SReT's hierarchical pyramid or other architecture-specific properties. Evaluating \MergeOver on larger recursive models, non-recursive hierarchical backbones, and standard ViTs with ToMe is necessary to establish generalisability.

Second, the canonical and stage-wise schedules differ in parameterisation, depth indexing, and reduction trajectory. Since we do not evaluate matched per-stage fractional constant and linear schedules under equivalent reduction budgets, our comparisons evaluate the complete scheduling formulations and do not isolate schedule shape alone. Thus, they do not establish that the observed schedule behaviour is specific to recursive architectures.

GPU PAM is measured using the CUDA allocator, whereas CPU memory is measured as a change in process Resident Set Size ($\Delta$RSS). The latter includes allocator caching, backend workspaces, and physical page-residency behaviour and does not isolate activation memory. CPU memory results are therefore provided only as process-level observations and are not directly comparable to GPU PAM.

Furthermore, the \texttt{thop} FLOPs estimates do not fully capture BSM matching, token-mass tracking, sequence manipulation, or \emph{Unmerge} tracking. We also do not perform low-level profiling or optimisation. Consequently, the causes of the observed platform- and batch-dependent performance differences cannot be isolated.

Finally, \MergeOver is evaluated relative to an uncompressed recursive baseline and does not establish superiority over conventional efficient ViTs. The modest Raspberry Pi latency improvement at batch size 1 demonstrates that practical single-stream edge deployment remains challenging. Future work could combine \MergeOver with low-level kernel optimisation and complementary post-training techniques, such as quantisation.

\section{Conclusion}
\label{sec:conclusion}
We presented \MergeOver and demonstrated the feasibility of combining the recursive weight-sharing of SReT with the dynamic token merging of ToMe as a post-training, multi-axis compression approach. By resolving the structural incompatibilities between hierarchical recursive architectures and dynamic sequence reduction, our approach enables token merging to be applied without retraining. Benchmarked on ImageNet-1K, our selected stage-wise single-shot configuration reduces top-1 accuracy by only 1.47 percentage points, which is negligible in many use-cases.

\emph{Considering the results from \Cref{sec:results}, the top performer scheduling strategy configuration, in terms of improved throughput, PAM and latency, while preserving the model accuracy, is the $\rho_{\mathrm{shot}} = 0.25$. The achieved throughput and PAM improvements for batch size 16 are 21.7\% increase and 38.4\% decrease on the GPU platform. Focusing on the latency improvements for x86 and ARM CPU variants, we observe 30.0\% reduction and 17.6\% reduction, respectively, similarly for batch size 16.} 

These results demonstrate that the hardware benefits of token reduction depend strongly on the processing platform and batch size. While there can be setups with minimal improvements, certain batch sizes, e.g., 16, show immensely better batch processing performance. Though further algorithmic and experimental refinements are warranted, \MergeOver provides a post-training baseline for integrating token merging into hierarchical recursive Transformers and establishes a foundation for further low-level optimisation and evaluation, especially on resource-constrained hardware.



\printbibliography

\appendix

\section{Extended results}
\label{app:extended-results}
This appendix reports the complete batch-dependent measurements underlying the hardware-efficiency analysis. For each platform, we compare the recursive ViT baseline with the evaluated token-reduction schedules across all supported batch sizes. The tables report Latency (Lat), Throughput (ThP), Peak Activation Memory (PAM), and process Resident Set Size ($\Delta$RSS) where applicable. Percentage changes shown next to the measurements are computed relative to the recursive ViT baseline.

\Cref{tab:gpu-batch-dependent-results} presents the GPU measurements, while \Cref{tab:cpu-batch-dependent-results,tab:pi-batch-dependent-results} report the corresponding results for the x86 and ARM CPU platforms, respectively.

\begin{table*}[htbp]
\centering
\setlength{\tabcolsep}{4pt}
\caption{GPU performance on ImageNet-1K across all tested batch sizes (1--128). We report Throughput (ThP) and Peak Activation Memory (PAM) measured on an NVIDIA GeForce RTX 4060 Ti. Our selected configuration is highlighted. Percentage changes are relative to the SReT baseline.}
\label{tab:gpu-batch-dependent-results}
\begin{tabular}{lllllllll}
\toprule
& \multicolumn{8}{c}{\textbf{Batch size}} \\
\cmidrule(lr){2-9}
\textbf{Metric} &
\textbf{1} &
\textbf{2} &
\textbf{4} &
\textbf{8} &
\textbf{16} &
\textbf{32} &
\textbf{64} &
\textbf{128} \\
\midrule

\multicolumn{9}{c}{\textbf{Recursive ViT -- SReT-Tiny-Distill~\cite{sret-shen2022}}} \\
\addlinespace[4pt]
\rowHLBaseline
\textbf{ThP (img/s)}
& 223.83 & 438.91 & 849.45 & 1213.81
& 1304.58 & 1256.95 & 1174.00 & 1076.17 \\
\rowHLBaseline
\textbf{PAM (MB)}
& 6.22 & 12.43 & 24.87 & 49.73
& 99.47 & 199.91 & 397.88 & 795.76 \\
\midrule

\multicolumn{9}{c}{\textit{Constant reduction: $r_{\mathrm{const}}=10$}} \\
\addlinespace[4pt]
\textbf{ThP (img/s)}
& 110.24\negR{50.7}
& 221.19\negR{49.6}
& 435.09\negR{48.8}
& 772.07\negR{36.4}
& 1163.09\negR{10.8}
& 1324.10\posG{5.3}
& 1275.88\posG{8.7}
& 1180.51\posG{9.7} \\
\textbf{PAM (MB)}
& 6.01\negG{3.3}
& 12.03\negG{3.2}
& 24.06\negG{3.2}
& 48.30\negG{2.9}
& 96.50\negG{3.0}
& 192.51\negG{3.7}
& 385.02\negG{3.2}
& 769.79\negG{3.3} \\
\midrule

\multicolumn{9}{c}{\textit{Constant reduction: $r_{\mathrm{const}}=20$}} \\
\addlinespace[4pt]
\textbf{ThP (img/s)}
& 109.88\negR{50.9}
& 217.68\negR{50.4}
& 428.69\negR{49.5}
& 781.68\negR{35.6}
& 1177.86\negR{9.7}
& 1475.31\posG{17.4}
& 1439.29\posG{22.6}
& 1340.50\posG{24.6} \\
\textbf{PAM (MB)}
& 5.91\negG{4.9}
& 11.82\negG{4.9}
& 23.63\negG{5.0}
& 47.55\negG{4.4}
& 94.95\negG{4.5}
& 189.15\negG{5.4}
& 380.27\negG{4.4}
& 756.22\negG{5.0} \\
\midrule

\multicolumn{9}{c}{\textit{Linear reduction: $r_{\mathrm{lin}}=10$}} \\
\addlinespace[4pt]
\textbf{ThP (img/s)}
& 111.56\negR{50.2}
& 222.00\negR{49.4}
& 441.16\negR{48.1}
& 821.16\negR{32.3}
& 1218.04\negR{6.6}
& 1326.12\posG{5.5}
& 1275.77\posG{8.7}
& 1181.37\posG{9.8} \\
\textbf{PAM (MB)}
& 5.91\negG{4.9}
& 11.82\negG{4.9}
& 23.63\negG{5.0}
& 47.55\negG{4.4}
& 94.95\negG{4.5}
& 189.15\negG{5.4}
& 380.27\negG{4.4}
& 756.22\negG{5.0} \\
\midrule

\multicolumn{9}{c}{\textit{Linear reduction: $r_{\mathrm{lin}}=20$}} \\
\addlinespace[4pt]
\textbf{ThP (img/s)}
& 117.12\negR{47.7}
& 224.51\negR{48.8}
& 440.79\negR{48.1}
& 816.08\negR{32.8}
& 1227.17\negR{5.9}
& 1554.19\posG{23.6}
& 1540.13\posG{31.2}
& 1440.20\posG{33.8} \\
\textbf{PAM (MB)}
& 5.70\negG{8.3}
& 11.40\negG{8.3}
& 22.80\negG{8.3}
& 46.06\negG{7.4}
& 91.89\negG{7.6}
& 183.15\negG{8.4}
& 366.76\negG{7.8}
& 729.46\negG{8.3} \\
\midrule

\multicolumn{9}{c}{\textit{Exponential reduction:
$\rho_{\mathrm{exp}}=0.25,\ \alpha=0.3$}} \\
\addlinespace[4pt]
\textbf{ThP (img/s)}
& 135.80\negR{39.3}
& 268.45\negR{38.8}
& 534.41\negR{37.1}
& 1007.02\negR{17.0}
& 1597.22\posG{22.4}
& 1665.33\posG{32.5}
& 1626.84\posG{38.6}
& 1477.07\posG{37.3} \\
\textbf{PAM (MB)}
& 3.90\negG{37.3}
& 7.72\negG{37.9}
& 15.37\negG{38.2}
& 30.67\negG{38.3}
& 61.27\negG{38.4}
& 123.13\negG{38.4}
& 245.46\negG{38.3}
& 489.38\negG{38.5} \\
\midrule

\multicolumn{9}{c}{\textit{Exponential reduction:
$\rho_{\mathrm{exp}}=0.40,\ \alpha=0.3$}} \\
\addlinespace[4pt]
\textbf{ThP (img/s)}
& 136.99\negR{38.8}
& 271.92\negR{38.0}
& 543.60\negR{36.0}
& 1055.32\negR{13.1}
& 1770.14\posG{35.7}
& 2142.17\posG{70.4}
& 2146.38\posG{82.8}
& 2017.02\posG{87.4} \\
\textbf{PAM (MB)}
& 3.90\negG{37.3}
& 7.72\negG{37.9}
& 15.37\negG{38.2}
& 30.67\negG{38.3}
& 48.94\negG{50.8}
& 97.88\negG{51.0}
& 195.76\negG{50.8}
& 391.51\negG{50.8} \\
\midrule

\multicolumn{9}{c}{\textit{Single-shot reduction:
$\rho_{\mathrm{shot}}=0.25$}} \\
\addlinespace[4pt]
\rowHLBest
\textbf{ThP (img/s)}
& 175.33\negR{21.7}
& 344.65\negR{21.5}
& 686.66\negR{19.2}
& 1263.14\posG{4.1}
& 1588.00\posG{21.7}
& 1573.91\posG{25.2}
& 1491.64\posG{27.1}
& 1364.47\posG{26.8} \\
\rowHLBest
\textbf{PAM (MB)}
& 3.90\negG{37.3}
& 7.72\negG{37.9}
& 15.37\negG{38.2}
& 30.67\negG{38.3}
& 61.27\negG{38.4}
& 123.13\negG{38.4}
& 245.46\negG{38.3}
& 489.38\negG{38.5} \\
\midrule

\multicolumn{9}{c}{\textit{Single-shot reduction:
$\rho_{\mathrm{shot}}=0.40$}} \\
\addlinespace[4pt]
\textbf{ThP (img/s)}
& 174.42\negR{22.1}
& 345.86\negR{21.2}
& 690.04\negR{18.8}
& 1324.94\posG{9.2}
& 1985.57\posG{52.2}
& 1944.38\posG{54.7}
& 1883.19\posG{60.4}
& 1716.86\posG{59.5} \\
\textbf{PAM (MB)}
& 3.90\negG{37.3}
& 7.72\negG{37.9}
& 15.37\negG{38.2}
& 30.67\negG{38.3}
& 48.94\negG{50.8}
& 97.88\negG{51.0}
& 195.76\negG{50.8}
& 391.51\negG{50.8} \\
\bottomrule
\end{tabular}
\end{table*}

\begin{table*}[htbp]
\centering
\caption{x86 CPU performance on ImageNet-1K across all tested batch sizes (1--16). We report Latency (Lat), Throughput (ThP), and change in process Resident Set Size ($\Delta$RSS) measured on an Intel Core Ultra 9 285K using four threads. Our selected configuration is highlighted. Percentage changes are relative to the SReT baseline.}
\label{tab:cpu-batch-dependent-results}
\begin{tabular}{llllll}
\toprule
& \multicolumn{5}{c}{\textbf{Batch size}} \\
\cmidrule(lr){2-6}
\textbf{Metric} &
\textbf{1} &
\textbf{2} &
\textbf{4} &
\textbf{8} &
\textbf{16} \\
\midrule

\multicolumn{6}{c}{\textbf{Recursive ViT -- SReT-Tiny-Distill~\cite{sret-shen2022}}} \\
\addlinespace[4pt]
\rowHLBaseline
\textbf{Lat (ms)}
& 14.94 & 22.76 & 36.26 & 68.28 & 144.32 \\
\rowHLBaseline
\textbf{ThP (img/s)}
& 66.93 & 87.86 & 110.30 & 117.17 & 110.86 \\
\rowHLBaseline
\textbf{$\Delta$RSS (MB)}
& 19.94 & 25.14 & 38.70 & 57.67 & 92.97 \\
\midrule

\multicolumn{6}{c}{\textit{Constant reduction:
$r_{\mathrm{const}}=10$}} \\
\addlinespace[4pt]
\textbf{Lat (ms)}
& 15.75\posR{5.4}
& 24.23\posR{6.5}
& 38.21\posR{5.4}
& 61.36\negG{10.1}
& 122.95\negG{14.8} \\
\textbf{ThP (img/s)}
& 63.50\negR{5.1}
& 82.55\negR{6.0}
& 104.67\negR{5.1}
& 130.38\posG{11.3}
& 130.14\posG{17.4} \\
\textbf{$\Delta$RSS (MB)}
& 19.35\negG{3.0}
& 23.98\negG{4.6}
& 36.36\negG{6.0}
& 59.36\posR{2.9}
& 91.97\negG{1.1} \\
\midrule

\multicolumn{6}{c}{\textit{Constant reduction:
$r_{\mathrm{const}}=20$}} \\
\addlinespace[4pt]
\textbf{Lat (ms)}
& 14.92\negG{0.1}
& 22.12\negG{2.8}
& 34.72\negG{4.2}
& 53.26\negG{22.0}
& 109.90\negG{23.8} \\
\textbf{ThP (img/s)}
& 67.01\posG{0.1}
& 90.43\posG{2.9}
& 115.21\posG{4.5}
& 150.22\posG{28.2}
& 145.59\posG{31.3} \\
\textbf{$\Delta$RSS (MB)}
& 19.77\negG{0.9}
& 21.65\negG{13.9}
& 36.05\negG{6.8}
& 58.09\posR{0.7}
& 90.08\negG{3.1} \\
\midrule

\multicolumn{6}{c}{\textit{Linear reduction:
$r_{\mathrm{lin}}=10$}} \\
\addlinespace[4pt]
\textbf{Lat (ms)}
& 16.08\posR{7.6}
& 24.88\posR{9.3}
& 37.74\posR{4.1}
& 61.37\negG{10.1}
& 124.76\negG{13.6} \\
\textbf{ThP (img/s)}
& 62.17\negR{7.1}
& 80.39\negR{8.5}
& 106.00\negR{3.9}
& 130.36\posG{11.3}
& 128.25\posG{15.7} \\
\textbf{$\Delta$RSS (MB)}
& 22.60\posR{13.3}
& 24.18\negG{3.8}
& 36.27\negG{6.3}
& 58.72\posR{1.8}
& 90.41\negG{2.8} \\
\midrule

\multicolumn{6}{c}{\textit{Linear reduction:
$r_{\mathrm{lin}}=20$}} \\
\addlinespace[4pt]
\textbf{Lat (ms)}
& 14.47\negG{3.1}
& 20.35\negG{10.6}
& 31.69\negG{12.6}
& 54.07\negG{20.8}
& 104.43\negG{27.6} \\
\textbf{ThP (img/s)}
& 69.10\posG{3.2}
& 98.28\posG{11.9}
& 126.21\posG{14.4}
& 147.96\posG{26.3}
& 153.21\posG{38.2} \\
\textbf{$\Delta$RSS (MB)}
& 22.07\posR{10.7}
& 24.04\negG{4.4}
& 33.59\negG{13.2}
& 56.53\negG{2.0}
& 79.02\negG{15.0} \\
\midrule

\multicolumn{6}{c}{\textit{Exponential reduction:
$\rho_{\mathrm{exp}}=0.25,\ \alpha=0.3$}} \\
\addlinespace[4pt]
\textbf{Lat (ms)}
& 12.91\negG{13.6}
& 20.02\negG{12.0}
& 32.65\negG{10.0}
& 53.52\negG{21.6}
& 102.65\negG{28.9} \\
\textbf{ThP (img/s)}
& 77.44\posG{15.7}
& 99.88\posG{13.7}
& 122.52\posG{11.1}
& 149.46\posG{27.6}
& 155.88\posG{40.6} \\
\textbf{$\Delta$RSS (MB)}
& 21.28\posR{6.7}
& 24.04\negG{4.4}
& 42.20\posR{9.0}
& 67.15\posR{16.4}
& 109.37\posR{17.6} \\
\midrule

\multicolumn{6}{c}{\textit{Exponential reduction:
$\rho_{\mathrm{exp}}=0.40,\ \alpha=0.3$}} \\
\addlinespace[4pt]
\textbf{Lat (ms)}
& 12.18\negG{18.5}
& 17.54\negG{22.9}
& 25.50\negG{29.7}
& 39.00\negG{42.9}
& 71.34\negG{50.6} \\
\textbf{ThP (img/s)}
& 82.10\posG{22.7}
& 114.06\posG{29.8}
& 156.87\posG{42.2}
& 205.14\posG{75.1}
& 224.26\posG{102.3} \\
\textbf{$\Delta$RSS (MB)}
& 21.13\posR{6.0}
& 20.70\negG{17.7}
& 40.61\posR{4.9}
& 62.31\posR{8.0}
& 99.77\posR{7.3} \\
\midrule

\multicolumn{6}{c}{\textit{Single-shot reduction:
$\rho_{\mathrm{shot}}=0.25$}} \\
\addlinespace[4pt]
\rowHLBest
\textbf{Lat (ms)}
& 12.27\negG{17.9}
& 21.20\negG{6.9}
& 30.75\negG{15.2}
& 54.13\negG{20.7}
& 100.98\negG{30.0} \\
\rowHLBest
\textbf{ThP (img/s)}
& 81.49\posG{21.8}
& 94.32\posG{7.4}
& 130.07\posG{17.9}
& 147.79\posG{26.1}
& 158.45\posG{42.9} \\
\rowHLBest
\textbf{$\Delta$RSS (MB)}
& 22.32\posR{11.9}
& 25.69\posR{2.2}
& 32.87\negG{15.1}
& 48.45\negG{16.0}
& 72.52\negG{22.0} \\
\midrule

\multicolumn{6}{c}{\textit{Single-shot reduction:
$\rho_{\mathrm{shot}}=0.40$}} \\
\addlinespace[4pt]
\textbf{Lat (ms)}
& 11.51\negG{23.0}
& 17.54\negG{22.9}
& 25.01\negG{31.0}
& 41.58\negG{39.1}
& 77.43\negG{46.3} \\
\textbf{ThP (img/s)}
& 86.84\posG{29.7}
& 114.04\posG{29.8}
& 159.94\posG{45.0}
& 192.42\posG{64.2}
& 206.63\posG{86.4} \\
\textbf{$\Delta$RSS (MB)}
& 21.13\posR{6.0}
& 22.73\negG{9.6}
& 40.60\posR{4.9}
& 62.28\posR{8.0}
& 99.52\posR{7.0} \\
\bottomrule
\end{tabular}
\end{table*}

\begin{table*}[htbp]
\centering
\caption{ARM CPU performance on ImageNet-1K across all tested batch
sizes (1--16). We report Latency (Lat),
Throughput (ThP), and change in process Resident Set Size
($\Delta$RSS) measured on a Raspberry Pi 5 with 16 GB of RAM.
Our selected configuration is highlighted. Percentage changes are
relative to the SReT baseline.}
\label{tab:pi-batch-dependent-results}
\begin{tabular}{llllll}
\toprule
& \multicolumn{5}{c}{\textbf{Batch size}} \\
\cmidrule(lr){2-6}
\textbf{Metric} &
\textbf{1} &
\textbf{2} &
\textbf{4} &
\textbf{8} &
\textbf{16} \\
\midrule

\multicolumn{6}{c}{\textbf{Recursive ViT -- SReT-Tiny-Distill~\cite{sret-shen2022}}} \\
\addlinespace[4pt]
\rowHLBaseline
\textbf{Lat (ms)}
& 142.82 & 204.02 & 346.97 & 637.68 & 1236.71 \\
\rowHLBaseline
\textbf{ThP (img/s)}
& 7.00 & 9.80 & 11.53 & 12.55 & 12.94 \\
\rowHLBaseline
\textbf{$\Delta$RSS (MB)}
& 21.89 & 28.23 & 50.91 & 82.64 & 163.23 \\
\midrule

\multicolumn{6}{c}{\textit{Constant reduction:
$r_{\mathrm{const}}=10$}} \\
\addlinespace[4pt]
\textbf{Lat (ms)}
& 159.61\posR{11.8}
& 214.50\posR{5.1}
& 351.50\posR{1.3}
& 620.03\negG{2.8}
& 1145.12\negG{7.4} \\
\textbf{ThP (img/s)}
& 6.27\negR{10.4}
& 9.32\negR{4.9}
& 11.38\negR{1.3}
& 12.90\posG{2.8}
& 13.97\posG{8.0} \\
\textbf{$\Delta$RSS (MB)}
& 24.86\posR{13.6}
& 31.41\posR{11.3}
& 53.61\posR{5.3}
& 85.42\posR{3.4}
& 135.69\negG{16.9} \\
\midrule

\multicolumn{6}{c}{\textit{Constant reduction:
$r_{\mathrm{const}}=20$}} \\
\addlinespace[4pt]
\textbf{Lat (ms)}
& 144.50\posR{1.2}
& 194.64\negG{4.6}
& 311.57\negG{10.2}
& 552.06\negG{13.4}
& 1011.14\negG{18.2} \\
\textbf{ThP (img/s)}
& 6.92\negR{1.1}
& 10.28\posG{4.9}
& 12.84\posG{11.4}
& 14.49\posG{15.5}
& 15.82\posG{22.3} \\
\textbf{$\Delta$RSS (MB)}
& 24.16\posR{10.4}
& 30.67\posR{8.6}
& 53.47\posR{5.0}
& 85.06\posR{2.9}
& 143.36\negG{12.2} \\
\midrule

\multicolumn{6}{c}{\textit{Linear reduction:
$r_{\mathrm{lin}}=10$}} \\
\addlinespace[4pt]
\textbf{Lat (ms)}
& 159.41\posR{11.6}
& 216.47\posR{6.1}
& 353.83\posR{2.0}
& 616.37\negG{3.3}
& 1167.84\negG{5.6} \\
\textbf{ThP (img/s)}
& 6.27\negR{10.4}
& 9.24\negR{5.7}
& 11.30\negR{2.0}
& 12.98\posG{3.4}
& 13.70\posG{5.9} \\
\textbf{$\Delta$RSS (MB)}
& 23.88\posR{9.1}
& 30.50\posR{8.0}
& 53.05\posR{4.2}
& 84.86\posR{2.7}
& 142.53\negG{12.7} \\
\midrule

\multicolumn{6}{c}{\textit{Linear reduction:
$r_{\mathrm{lin}}=20$}} \\
\addlinespace[4pt]
\textbf{Lat (ms)}
& 140.90\negG{1.3}
& 184.37\negG{9.6}
& 293.57\negG{15.4}
& 512.92\negG{19.6}
& 965.58\negG{21.9} \\
\textbf{ThP (img/s)}
& 7.10\posG{1.4}
& 10.85\posG{10.7}
& 13.63\posG{18.2}
& 15.60\posG{24.3}
& 16.57\posG{28.1} \\
\textbf{$\Delta$RSS (MB)}
& 23.97\posR{9.5}
& 30.53\posR{8.1}
& 52.98\posR{4.1}
& 89.22\posR{8.0}
& 136.25\negG{16.5} \\
\midrule

\multicolumn{6}{c}{\textit{Exponential reduction:
$\rho_{\mathrm{exp}}=0.25,\ \alpha=0.3$}} \\
\addlinespace[4pt]
\textbf{Lat (ms)}
& 140.94\negG{1.3}
& 186.13\negG{8.8}
& 298.38\negG{14.0}
& 526.78\negG{17.4}
& 975.44\negG{21.1} \\
\textbf{ThP (img/s)}
& 7.09\posG{1.3}
& 10.75\posG{9.7}
& 13.41\posG{16.3}
& 15.19\posG{21.0}
& 16.40\posG{26.7} \\
\textbf{$\Delta$RSS (MB)}
& 22.83\posR{4.3}
& 27.81\negG{1.5}
& 43.38\negG{14.8}
& 66.62\negG{19.4}
& 85.98\negG{47.3} \\
\midrule

\multicolumn{6}{c}{\textit{Exponential reduction:
$\rho_{\mathrm{exp}}=0.40,\ \alpha=0.3$}} \\
\addlinespace[4pt]
\textbf{Lat (ms)}
& 126.84\negG{11.2}
& 165.14\negG{19.1}
& 256.52\negG{26.1}
& 421.88\negG{33.8}
& 774.81\negG{37.3} \\
\textbf{ThP (img/s)}
& 7.88\posG{12.6}
& 12.11\posG{23.6}
& 15.59\posG{35.2}
& 18.96\posG{51.1}
& 20.65\posG{59.6} \\
\textbf{$\Delta$RSS (MB)}
& 22.73\posR{3.8}
& 27.70\negG{1.9}
& 43.17\negG{15.2}
& 66.45\negG{19.6}
& 61.89\negG{62.1} \\
\midrule

\multicolumn{6}{c}{\textit{Single-shot reduction:
$\rho_{\mathrm{shot}}=0.25$}} \\
\addlinespace[4pt]
\rowHLBest
\textbf{Lat (ms)}
& 139.42\negG{2.4}
& 187.30\negG{8.2}
& 295.35\negG{14.9}
& 539.22\negG{15.4}
& 1018.61\negG{17.6} \\
\rowHLBest
\textbf{ThP (img/s)}
& 7.17\posG{2.4}
& 10.68\posG{9.0}
& 13.54\posG{17.4}
& 14.84\posG{18.2}
& 15.71\posG{21.4} \\
\rowHLBest
\textbf{$\Delta$RSS (MB)}
& 22.75\posR{3.9}
& 27.70\negG{1.9}
& 43.36\negG{14.8}
& 66.20\negG{19.9}
& 98.66\negG{39.6} \\
\midrule

\multicolumn{6}{c}{\textit{Single-shot reduction:
$\rho_{\mathrm{shot}}=0.40$}} \\
\addlinespace[4pt]
\textbf{Lat (ms)}
& 129.49\negG{9.3}
& 166.75\negG{18.3}
& 267.28\negG{23.0}
& 460.08\negG{27.9}
& 860.18\negG{30.4} \\
\textbf{ThP (img/s)}
& 7.72\posG{10.3}
& 11.99\posG{22.3}
& 14.97\posG{29.8}
& 17.39\posG{38.6}
& 18.60\posG{43.7} \\
\textbf{$\Delta$RSS (MB)}
& 22.23\posR{1.6}
& 27.50\negG{2.6}
& 43.06\negG{15.4}
& 66.38\negG{19.7}
& 64.33\negG{60.6} \\
\bottomrule
\end{tabular}
\end{table*}


\end{document}